\documentclass{article}
\usepackage{spconf,amsmath,amssymb,graphicx,hyperref}

\usepackage{xcolor}
\usepackage{multirow}
\usepackage{graphicx}
\usepackage{booktabs}
\usepackage{subfloat}
\usepackage{bbding}
\usepackage{makecell}
\usepackage{colortbl}
\usepackage{placeins}
\usepackage{etoolbox}
\apptocmd{\thebibliography}{\setlength{\itemsep}{2pt}}{}{}

\title{Frequency-Domain Refinement with\\Multiscale Diffusion for Super Resolution}

\name{Xingjian Wang$^{\dagger}$,\quad Li Chai$^{\dagger}\sthanks{Corresponding author.}$,\quad Jiming Chen$^{\dagger}$}
\address{$^{\dagger}$Zhejiang University, College of Control Science and Engineering, Hangzhou, China}

\begin{document}

\maketitle

\begin{abstract}
The performance of single image super-resolution depends heavily on how to generate and complement high-frequency details to low-resolution images.
Recently, diffusion-based DDPM models exhibit great potential in generating high-quality details for super-resolution tasks.
They tend to directly predict high-frequency information of wide bandwidth by solely utilizing the high-resolution ground truth as the target for all sampling timesteps.
However, as a result, they encounter hallucination problem that they generate mismatching artifacts.
To tackle this problem and achieve higher-quality super-resolution, we propose a novel Frequency Domain-guided multiscale Diffusion model (FDDiff), which decomposes the high-frequency information complementing process into finer-grained steps.
In particular, a wavelet packet-based frequency degradation pyramid is developed to provide multiscale intermediate targets with increasing bandwidth.
Based on these targets, FDDiff guides reverse diffusion process to progressively complement missing high-frequency details over timesteps.
Moreover, a multiscale frequency refinement network is designed to predict the required high-frequency components at multiple scales within one unified network.
Comprehensive evaluations on popular benchmarks are conducted, and demonstrate that FDDiff outperforms prior generative methods with higher-fidelity super-resolution results.
\end{abstract}


\section{Introduction}
\label{sec:intro}

Super-resolution (SR) for single image is a crucial task, playing a vital role in enhancing the quality of low-resolution (LR) images for various downstream tasks.
The natural or artificial degradation process of LR images can be regarded as the union of low-pass filtering and noise.
The main difficulty of SR lies in the restoration of missing high-frequency details.

\begin{figure}[t]
    \centering
    \includegraphics[width=0.45\textwidth]{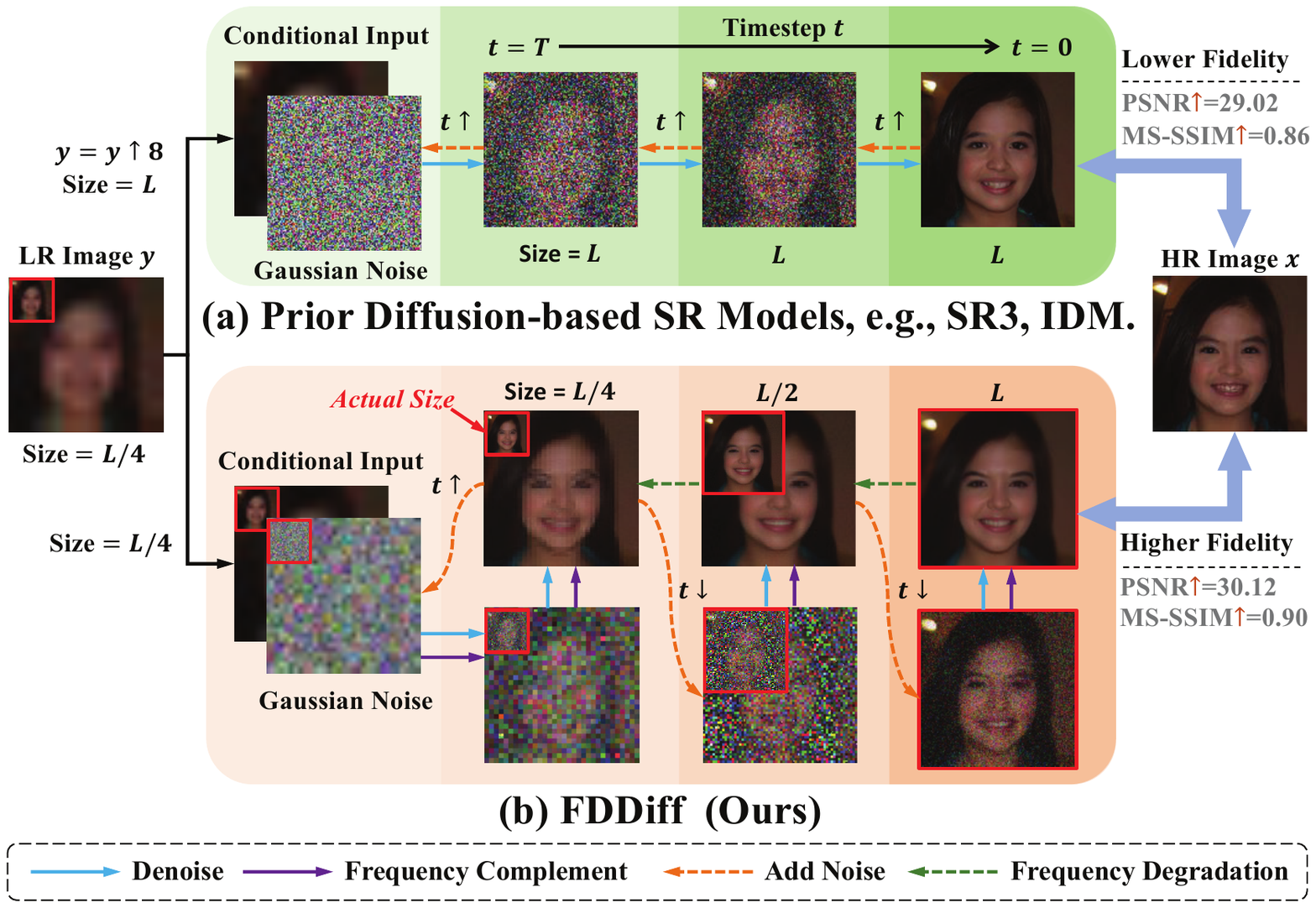}
    \vspace{-1em}
    \caption{Difference between prior diffusion-based SR models\cite{SR32023,srdiff2022,IDM2023,DiWa2023} and FDDiff. $4\times$ SR is taken as example, and images in red boxes represent the actual size. Benefiting from multiscale frequency refinement, FDDiff generates higher-fidelity SR results consistent with ground truth.}
    \vspace{-1em}
    \label{Fig.Intro}
\end{figure}

Recent SR methods can be split into two categories, namely regression-based methods and generative methods.
Among them, regression-based methods \cite{EDSR2017,LIIF2021} directly learn the LR-to-HR mapping.
Although achieving relatively small pixel-wise errors, regression-based methods suffer from low perceptual quality with insufficient high-frequency details.
Instead, generative methods \cite{SR32023,srdiff2022,IDM2023,DiWa2023,esrgan2018,ranksrgan2019,HCFLow2021,lar2022} have seen success in generating realistic details, which leverage the prior distribution learned from training dataset for super-resolution.
However, they also have various limitations.
For instance, GAN-based SR methods \cite{esrgan2018,ranksrgan2019} are prone to unstable training and mode collapse, while Normalizing Flow-based \cite{HCFLow2021} and VAE-based \cite{lar2022} SR methods suffer from less visually pleasing SR results.
Recently, diffusion-based SR methods \cite{SR32023,srdiff2022,IDM2023,DiWa2023} using DDPM \cite{DDPM2020} have made encouraging progress in high-quality generative SR.

\begin{figure*}[t]
  \centering
  \includegraphics[width=0.9\textwidth]{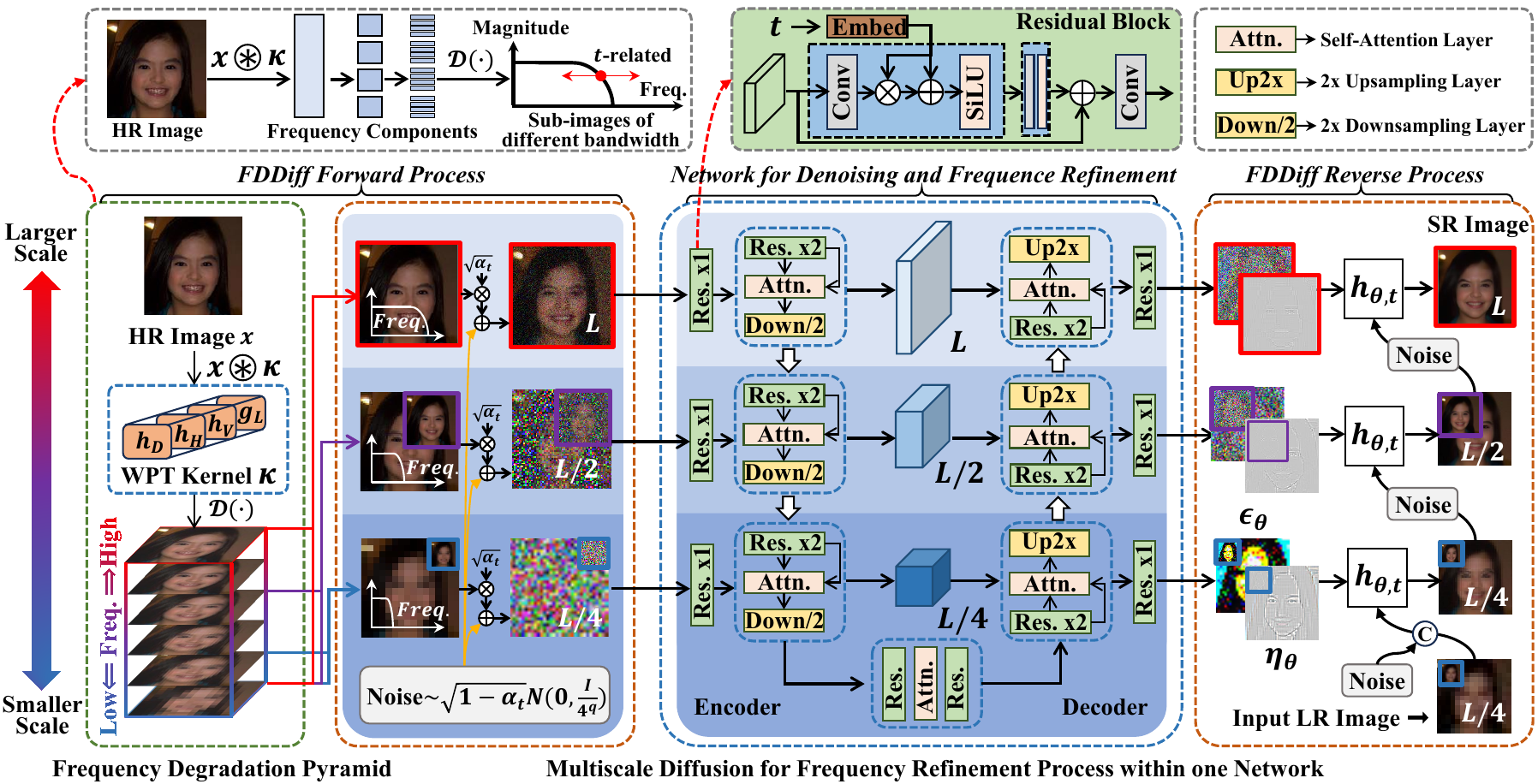}
  \vspace{-1em}
  \caption{Overview framework of FDDiff, which is tailored to progressively complement the missing frequency components to LR images at different scales. Upscaling factor 4 is taken as example for presentation. }
  \vspace{-1em}
  \label{Fig.Framework}
\end{figure*}

Although encouraging progress have been made by these diffusion-based SR methods, they suffer from hallucination problem \cite{Hallucination}, limiting their practical deployments like face SR.
That is, they generate hallucinated details which fail to perceptually match the given LR images, especially for large upscaling factors.
The reason lies in the fact that they aim to directly predict high-frequency information within a relatively wide bandwidth all at once, setting HR ground truth as sole target throughout the whole reverse diffusion process.
As illustrated in DDIM \cite{DDIM2022}, the diffusion process can be a non-Markovian process as long as the latent conditional distribution at time $t$ conditioned on its initial state remains the same with the one in DDPM \cite{DDPM2020}.
Therefore, it is recommended to construct a non-Markovian process to provide prior frequency-domain guidance for higher-fidelity SR.

To address abovementioned problem, we propose a novel Frequency Domain-guided multiscale Diffusion model (FDDiff) to progressively complement the missing high-frequency components in LR images.
Avoiding the difficulties of the direct transition from a simple Gaussian distribution to the complex HR image distribution in prior works \cite{SR32023,srdiff2022,IDM2023,DiWa2023}, we propose a wavelet packet-based frequency degradation pyramid to additionally introduce easily accessible distributions to the diffusion reverse process.
These intermediate distributions can be regard as temporally varying intermediate targets with progressively refined high-frequency information and hierachical scales.
Based on the frequency degradation pyramid, the corresponding diffusion process for multiscale frequency refinement is proposed to progressively predict the missing high-frequency information within a small local bandwidth step by steps.
Leveraging prior frequency-domain guidance, the multiscale frequency refinement will generate more faithful details to the ground truth.
We evaluate FDDiff on various super-resolution tasks from face images \cite{FFHQ2019,CelebAHQ2018} to general images \cite{DIV2K2017}. 
Experimental results demonstrate that FDDiff achieve higher fidelity than other concurrent works.

\section{Method\label{sec:method}}
\noindent
\textbf{Overview.}
As illustrated in Fig. \ref{Fig.Framework}, we propose a multiscale frequency refinement framework named FDDiff based on non-Markovian diffusion process.
FDDiff is designed to progressively generates missing high-frequency details for LR images, aiming at higher-fidelity SR with details consistent to the LR input.
The diffusion forward process of FDDiff corrupts an image by progressively adding noise and removing high-frequency components of specific bandwidths, thereby simulating the degradation process into LR images.
In this way, the introduced prior constrains the solution space to force the model to reverse a predefined degradation path.
The corresponding reverse process denoises and predicts missing details at different scales simultaneously within one well-designed multiscale network.
To provide  prior guidance for frequency refinement process, we design a frequency degradation pyramid based on wavelet packet.
Then we assign these sub-images to different timesteps through non-linear interpolation to provide high-frequency intermediate targets.

\noindent
\textbf{Wavelet Packet-based Frequency Degradation Pyramid.}
Let $\left\{\boldsymbol{x}, \boldsymbol{y}\right\}$ be a given HR-LR image pair, where $\boldsymbol{x}\in \mathbb{R}^{3\times H\times W}$ represents the expected high-resolution output images, $\boldsymbol{y}\in \mathbb{R}^{3\times \frac{H}{2^p}\times \frac{W}{2^p}}$ represents low-resolution images, and $2^p$ denotes upscaling factor.
Multiscale frequency degradation pyramid is designed to obtain the temporally degraded counterpart $\hat{\boldsymbol{x}}_t$ conditioned on discrete timestep $t\in[1,T]$.

Firstly, leveraging decomposition process $\mathcal{F}$ of 2D Haar Wavelet Packet Transform (2D-WPT), $\boldsymbol{x}$ is iteratively convolved $p$ times by four Haar kernels $\mathcal{K}=\{\boldsymbol{h}_{D}, \boldsymbol{h}_{H}, \boldsymbol{h}_{V}, \boldsymbol{g}_{L}\}$ with stride$=$2, projected onto $s^\textrm{th}$-stage frequency domain.
This results in $\boldsymbol{x}_p=\mathcal{F}(\boldsymbol{x}, p)\in \mathbb{R}^{(3\cdot4^p)\times \frac{H}{2^p}\times W\frac{H}{2^p}}$ with increasing channel dimension and decreasing spatial resolution.
$\boldsymbol{x}_p$ can be split into $4^p$ parts of different frequency coefficients, i.e., $\boldsymbol{x}_p = [\boldsymbol{x}_p^{4^p},\boldsymbol{x}_p^{4^p-1},...,\boldsymbol{x}_p^2,\boldsymbol{x}_p^1]$ with frequency descending along the index.
Low-pass filter $\mathcal{D}(\cdot,j)$ is defined to obtain top-j low-frequency components from $\boldsymbol{x}_p$, i.e., $\mathcal{D}(\boldsymbol{x}_p,j)=[\boldsymbol{x}_p^j,...,\boldsymbol{x}_p^2,\boldsymbol{x}_p^1]$.
Then we utilize the restoration process $\mathcal{F}^{-1}$ of 2D-WPT to restore $\mathcal{D}(\boldsymbol{x}_p,j)$ with $\lceil\mathrm{log}_4(j)\rceil$ times from frequency domain into image domain, i.e., $\tilde{\boldsymbol{x}}_{p,j} =\mathcal{F}^{-1}\left(\mathcal{D}(\boldsymbol{x}_p,j), \lceil\mathrm{log}_4(j)\rceil\right)$ where $\lceil\cdot\rceil$ represents the ceiling function.
Opposite to $2^p$ upscaling SR, the pyramid degrades $\tilde{\boldsymbol{x}}_{p,4^p}=\boldsymbol{x}$ to LR image $\tilde{\boldsymbol{x}}_{p,1}=\boldsymbol{x}_p^1\in \mathbb{R}^{3\times \frac{H}{2^p}\times \frac{W}{2^p}}$.
In this way, it creates $4^p$ states of $x$ with different frequency bandwidths and scales.

Furthermore, we assign aforementioned $4^p$ states of $x$ to $t\!\in\![1,T]$ timesteps.
Since the low-frequency components retaining larger energy are harder to be restored, we sample more timesteps for lower index $j$ by power function $j = \lceil 4^p\left(1-\frac{(T-t)^2}{T^2}\right)\rceil$.
Then frequency degradation pyramid over $T$ timesteps can be derived from as follows,
\begin{equation}
\small
\hat{\boldsymbol{x}}_t \!\triangleq\!
\begin{cases}
    \frac{(t-t_{j-1})\tilde{\boldsymbol{x}}_{p,j}+(t_{j}-t)\tilde{\boldsymbol{x}}_{p,j-1}}{t_{j}-t_{j-1}}\!&\!t_{j-1}\!<\!t\!\le\!t_{j}, 2\!\le \!j\!<\!4^p,\\
    \frac{(t-t_{j})\boldsymbol{y}+(T-t)\tilde{\boldsymbol{x}}_{p,4^p}}{T-t_{j}}\!&\!t_{j}< t \le T, j=4^p,
\end{cases}
\label{eq.hat_x}
\end{equation}
where the key timestep $t_j$ corresponding to 2x downscaling transition from $\frac{1}{2^{t_j-1}}$ to $\frac{1}{2^{t_j}}$ is set to $t_j = \lfloor T-T\sqrt{1-\frac{j-1}{4^p}}\rfloor$.
As for the boundary case, $\hat{\boldsymbol{x}}_1 = x$ and $\hat{\boldsymbol{x}}_T = y$.

It is worth noting that $\hat{\boldsymbol{x}}_t$ undergoes scale changing from $t_j$ to $t_j-1$, directly corresponding to multiscale SR.
Therefore, the corresponding reverse process serves as a frequency complement chain from $\boldsymbol{y}$ to $\boldsymbol{x}$, providing missing high-frequency component $\boldsymbol{\eta}_t$ as additional targets for FDDiff,
\begin{equation}
    \boldsymbol{\eta}_t = \hat{\boldsymbol{x}}_{t_{j-1}} - \hat{\boldsymbol{x}}_t, \qquad t_{j-1}< t \le t_{j}.
\end{equation}

\noindent
\textbf{Multiscale Frequency Refinement Diffusion.}
Directly sampling around $\hat{\boldsymbol{x}}_t$ with Gaussian noise, we obtain noised latent variable $u_t$, whose conditional distribution $q(\boldsymbol{u}_t|\boldsymbol{u}_0)$ is
\begin{equation}
    q(\boldsymbol{u}_t|\hat{\boldsymbol{x}}_t,\boldsymbol{y})=\mathcal{N}(\boldsymbol{u}_t;\sqrt{\alpha_t}\hat{\boldsymbol{x}}_t,(1-\alpha_t)\boldsymbol{I}),
\label{eq.marginal}
\end{equation}
where noise hyperparameter $\alpha_t$ defined in DDPM \cite{DDPM2020} follows cosine schedule in \cite{DDIM2022}. 
$\alpha_0=1$ and $\alpha_t$ monotonically decreases to zero over time.
Evidently, $\boldsymbol{u}_1=\hat{\boldsymbol{x}}_1=\boldsymbol{x}$ when $t=1$.
To meet the marginal distribution, the corresponding non-Markovian diffusion forward process is given by,
\begin{equation}
    \resizebox{0.44\textwidth}{!}{$
    \boldsymbol{u}_t \!=\! \sqrt{\frac{\alpha_t}{\alpha_{t-1}}}\boldsymbol{u}_{t-1}\!\downarrow_{\gamma(t)}-\frac{\boldsymbol{\eta}_t\sqrt{\alpha_t}}{t_j-t_{j-1}}\!+\!\sqrt{\frac{1-\alpha_t}{1-\alpha_{t-1}}}\boldsymbol{\epsilon}_t,
    $}
\end{equation}
where $t_{j-1}< t \le t_{j}$ and $\boldsymbol{\epsilon}_t\sim\mathcal{N}(\boldsymbol{0},\boldsymbol{I})$ denotes the added Gaussian noise for every forward step.
For scale-changing timesteps, average downsampling operation $\downarrow_{\gamma(t)}$ with integer factor=$\gamma(t)$ is added.
That is, $\gamma(t)$ equals to $2$ if $t=t_j$ and $j=4^q,\;q\in\{0,1,2,...,p\}$, otherwise $\gamma(t)$ equals to $1$.
As a result, the variance of $\boldsymbol{\epsilon}_t$ decreases, which is sampled from $\mathcal{N}(\boldsymbol{0},\frac{1}{4^q}\boldsymbol{I})$ while $q=p-\lceil\mathrm{log}_4(4^p+1-j)\rceil$.

\noindent
\textbf{Reverse Process of Diffusion.}
For non-Markovian diffusion forward process, we design the reverse process based on DDIM \cite{DDIM2022} and f-DM \cite{fDM2023} to predict $\boldsymbol{u}_{t-1}$ from $\boldsymbol{u}_{t}$ within a multiscale network.
Specifically, the network with parameters $\theta$ is employed to estimate $\boldsymbol{\eta}_t$ with $\boldsymbol{\eta}_\theta$ and $\boldsymbol{\epsilon}_t$ with $\boldsymbol{\epsilon}_\theta$ simultaneously.
Consequently, the reverse process of multiscale frequency-refinement diffusion can be derived as,
\begin{equation}
\left\{\begin{aligned}
    &\boldsymbol{u}_{t-1} \!=\! \sqrt{\alpha_{t-1}}h_{\theta,t}(\boldsymbol{u}_t,\boldsymbol{\epsilon}_\theta,\boldsymbol{\eta}_\theta)\!\uparrow_{\gamma(t)}+\sqrt{1-\alpha_{t-1}}\boldsymbol{\epsilon}_{t-1},\\
    &h_{\theta,t}(\boldsymbol{u}_t,\boldsymbol{\epsilon}_\theta,\boldsymbol{\eta}_\theta) \!=\! \frac{\boldsymbol{u}_t-\sqrt{1-\alpha_t}\boldsymbol{\epsilon}_\theta}{\alpha_{t}}+\frac{\boldsymbol{\eta}_\theta}{t_j-t_{j-1}},\\
    &\boldsymbol{\epsilon}_\theta,\boldsymbol{\eta}_\theta = \mathrm{Net}_\theta(\boldsymbol{u}_t,\boldsymbol{y},t),
\end{aligned}\right.
\label{eq.reverse}
\end{equation}  
where $\mathrm{Net}_\theta^q(\cdot)$ denotes the network predicting $\boldsymbol{\epsilon}_\theta$ and $\boldsymbol{\eta}_\theta$ from $\boldsymbol{u}_t$ with conditional input $y$ and $t$. 
The $\mathrm{Net}_\theta^q(\cdot)$ has $p+1$ states for $p+1$ scales.
Given $t$, it refers to $q$-th state while $q=p-\lceil\mathrm{log}_4(4^p+1-j)\rceil$.
Function $h_{\theta,t}(\cdot)$ conducted by network is utilized to denoise and recover high-frequency components of the local narrow bandwidth from $\boldsymbol{x}_{t}$ to $\boldsymbol{x}_{t-1}$.
Symmetric to $\downarrow_{\gamma(t)}$ in forward process, upsampling operation $\uparrow_{\gamma(t)}$ with factor $\gamma(t)$ is also added to the reverse process.
According to $\mathcal{F}^{-1}$ of Haar wavelet, nearest upsampling is chosen for $\uparrow_{\gamma(t)}$.

\begin{table}
    \centering
    \caption{Quantitative comparison with generative models for $16\!\times \!16\!\rightarrow\!128\!\times\! 128$ face super-resolution on CelebA-HQ.}
    \resizebox{0.8\linewidth}{!}{
    \begin{tabular}{clcc}
        \toprule
        Categories & Methods & PSNR$\uparrow$ & SSIM$\uparrow$\\
        \midrule
        \multirow{2}{*}{GAN-based} & PULSE \cite{pulse2020} & 16.88 & 0.44 \\
        & FSRGAN \cite{fsrgan2018} & 23.01 & 0.62\\
        \midrule
        \multirow{5}{*}{Diffusion} & Regression \cite{SR32023} & 23.96 & 0.69\\
        & SR3 \cite{SR32023} & 23.04 & 0.65\\
        & DiWa \cite{DiWa2023} & 23.34 & 0.67 \\
        & IDM \cite{IDM2023} & 24.01 & \textbf{0.71}\\
        \midrule
        Diffusion & \textbf{FDDiff} & \textbf{24.52} & \textbf{0.71}\\
        \bottomrule
    \end{tabular}}
    \label{tab.8xfaceSR}
    \vspace{-1em}
\end{table}

\begin{table}
    \centering
    \caption{
        Quantitative comparison on DIV2K validation set of general images for $4\times$ natural super-resolution. \label{tab.4xDIV2K}
    }
    \resizebox{\linewidth}{!}{
    \begin{tabular}{clccc}
        \toprule 
        Categories & Methods & Datasets & PSNR $\uparrow$ & SSIM $\uparrow$ \\
        \hline
        \multirow{2}{*}{Regression} & EDSR \cite{EDSR2017} & DF2K & 28.98 & 0.83 \\
        & LIIF \cite{LIIF2021} & DF2K & 29.00 & 0.89 \\
        \hline \hline  
        VAE-based & LAR-SR \cite{lar2022} & DF2K & 27.03 & 0.77 \\
        \hline
        \multirow{2}{*}{GAN-based} & ESRGAN \cite{esrgan2018} & DF2K & 26.22 & 0.75 \\
        & RankSRGAN \cite{ranksrgan2019} & DF2K & 26.55 & 0.75 \\
        \hline 
        \multirow{1}{*}{Flow-based} & HCFlow \cite{HCFLow2021} & DF2K & 27.02 & 0.76 \\
        \hline 
        Flow+GAN & HCFlow++ \cite{HCFLow2021} & DF2K & 26.61 & 0.74 \\
        \hline 
        \multirow{2}{*}{Diffusion} 
        & SRDiff \cite{srdiff2022} & DIV2K & 27.41 & 0.79 \\
        & IDM \cite{IDM2023} & DIV2K & 27.10 & 0.77 \\
        \midrule
        Diffusion
            & \textbf{FDDiff} & DIV2K & \textbf{27.53} & \textbf{0.85} \\
        \bottomrule
    \end{tabular}
    }
    \vspace{-1em}
\end{table}

\noindent
\textbf{Training Objective.}
For timestep $t_{j-1}< t \le t_{j}$, generating the noiseless pre-state $\hat{\boldsymbol{x}}_{t_{j-1}}=\tilde{\boldsymbol{x}}_{j-1}$ from $\boldsymbol{u}_t$ is regarded as the training objective.
Since estimation of $\tilde{\boldsymbol{x}}_{j-1}$ can be derived as $\frac{\boldsymbol{u}_t-\sqrt{1-\alpha_t}\boldsymbol{\epsilon}_\theta}{\alpha_{t}}+\boldsymbol{\eta}_\theta$, loss function $\mathcal{L}(\theta)$ is defined as,
\begin{equation}
    \resizebox{0.45\textwidth}{!}{$
    \mathcal{L}(\theta)\!=\!\underset{(\mathbf{x}, \mathbf{y})}{\mathbb{E}}\underset{j}{\mathbb{E}}\!\left(\underset{t_{j-1}<t\le t_j}{\mathbb{E}}\frac{\alpha_t}{1-\alpha_t}\left\|\frac{\boldsymbol{u}_t-\sqrt{1-\alpha_t}\boldsymbol{\epsilon}_\theta}{\alpha_{t}}+\boldsymbol{\eta}_\theta-\tilde{\boldsymbol{x}}_{j-1}\right\|_1\right),
    $}
    \label{eq.loss}
\end{equation} 
where $\frac{\alpha_t}{1-\alpha_t}$ denotes the SNR weighting coefficients in \cite{DDIM2022}.

\noindent
\textbf{Diffusion Network Design.}
Given conditional input $\boldsymbol{y}$ and $t$, network $\mathrm{Net}_\theta^q(\cdot)$ in Eq. \ref{eq.reverse} is expected to denoise $\boldsymbol{\epsilon}_\theta$ and predict missing details $\boldsymbol{\eta}_\theta$ of different scales simultaneously.
Thus, it is designed with $p+1$ states for $2^p$ super resolution, following U-Net architecture in f-DM \cite{fDM2023}, consisting of $p+1$ encoders and decoders.
Specifically, for encoder/decoder, each block is composed of two residual blocks, a self-attention layer, and a $2\times$ downsampling/upsampling layer, except the last block.
As for conditional input, LR image $\boldsymbol{y}$ is concatenated with the input noised image at the channel dimension, and timestep $t$ is projected by linear layer to conduct learnable affine transformation in residual blocks.

\begin{figure}
    \centering
    \includegraphics[width=0.73\linewidth]{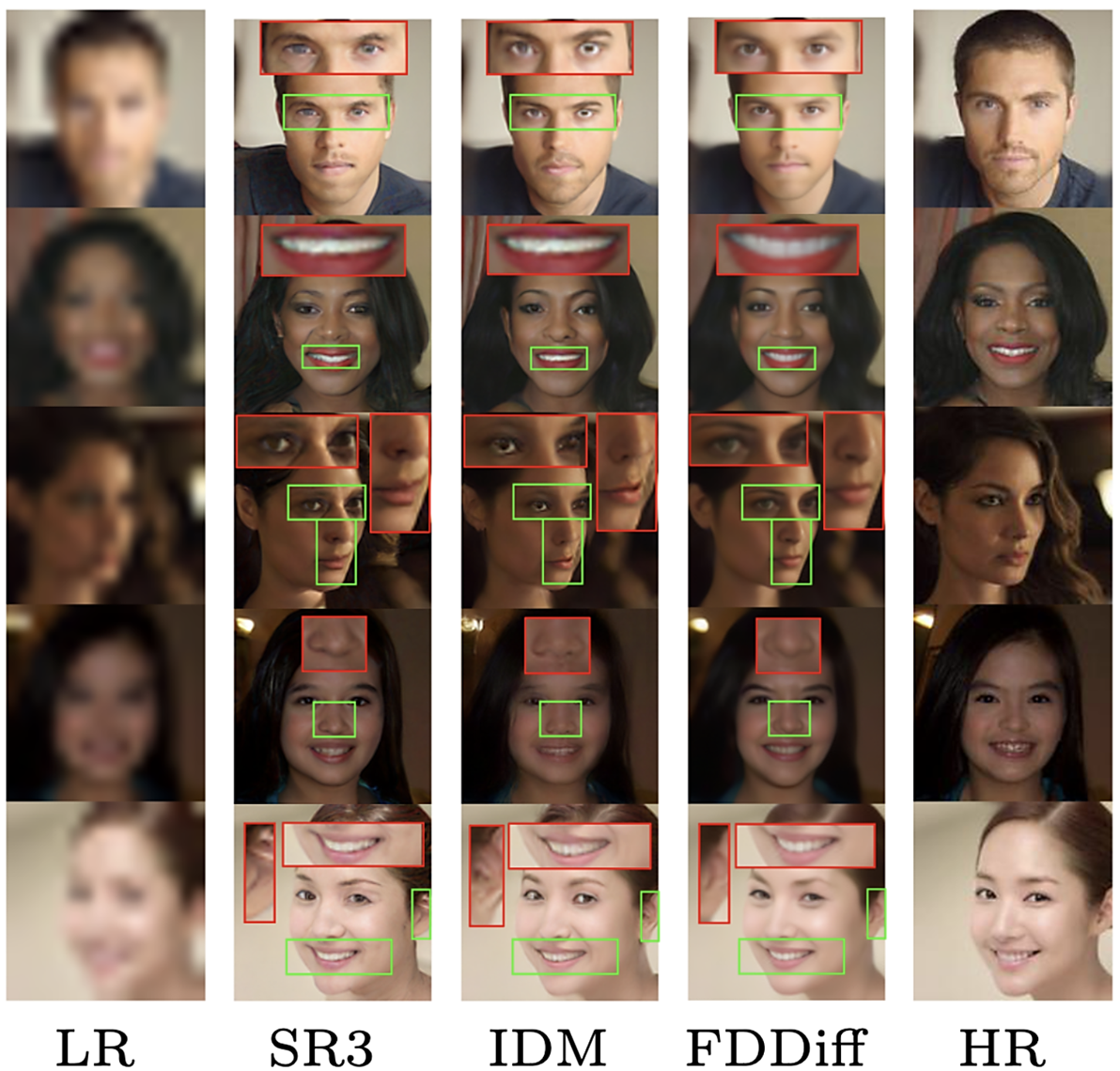}
    \vspace{-1em}
    \caption{Qualitative comparison on 8x SR for CelebA-HQ. For facial features in green boxes, their corresponding enlarged patches are shown in red boxes.}
    \vspace{-1em}
    \label{fig.face}
\end{figure}
\begin{table}
    \centering
    \caption{Real-world 4x SR comparison on RealSR-V3 \cite{realSR}.} 
    \label{tab.realsr}    
    \resizebox{0.7\linewidth}{!}{
    \begin{tabular}{lccc}
    \toprule
    Methods & PSNR & SSIM  & MUSIQ \\
    \midrule
    DASR \cite{DASR} & 27.02 & 0.7707 & 41.21 \\
    StableSR \cite{stableSR} & 24.65 & 0.7080 & \textbf{65.88} \\
    \midrule
    FDDiff & \textbf{26.13} & \textbf{0.7925} & 45.92 \\
    \bottomrule
    \end{tabular}
    }
    \vspace{-1em}
\end{table}
\begin{figure}
    \centering
    \includegraphics[width=0.8\linewidth]{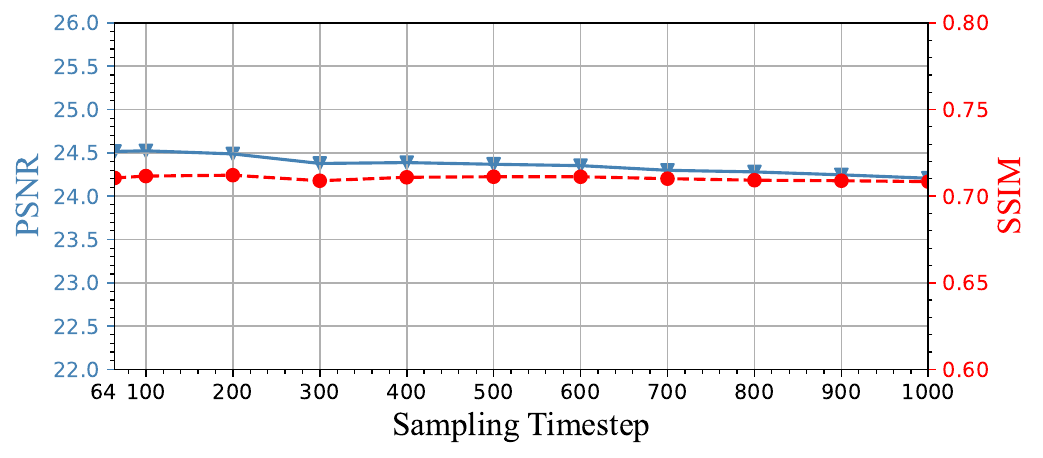}
    \vspace{-1em}
    \caption{Super-resoltion quality versus sampling timesteps.\label{Fig.ExpTimestep}}
    \vspace{-1em}
\end{figure}
\begin{table}
    \centering
    \caption{Ablation Study on different components of FDDiff. } 
    \label{tab.ExpAblationFDDiff}
    \resizebox{\linewidth}{!}{
    \begin{tabular}{c|cc|ccc|cc|cc}
    \Xhline{1px}
    \multirow{2}{*}{No.}&\multicolumn{2}{c|}{Scale Setting} & \multicolumn{3}{c|}{Target Number} & \multicolumn{2}{c|}{Target} & \multirow{2}{*}{PSNR} & \multirow{2}{*}{SSIM} \\
    \Xcline{2-3}{0.4pt}\Xcline{4-6}{0.4pt}\Xcline{7-8}{0.4pt}
    & Single & Multiple & Single & $p+1$ & $4^p$ &$\eta_t$ &$\epsilon_t$ & & \\
    \Xhline{1px}
    1&\Checkmark & \XSolidBrush & \Checkmark  & \XSolidBrush & \XSolidBrush & \Checkmark & \XSolidBrush & 24.09 & 0.7002 \\
    2&\XSolidBrush & \Checkmark & \XSolidBrush  & \Checkmark & \XSolidBrush & \Checkmark & \Checkmark & 24.48 & 0.7092 \\
    \rowcolor{gray!30}
    3&\XSolidBrush & \Checkmark & \XSolidBrush  & \XSolidBrush & \Checkmark & \Checkmark & \Checkmark & 24.52 & 0.7116 \\
    4&\XSolidBrush & \Checkmark & \XSolidBrush  & \XSolidBrush & \Checkmark & \Checkmark & \XSolidBrush & 21.87 & 0.6046 \\  
    \Xhline{1px}
    \end{tabular}}
    \vspace{-1em}
\end{table}

\section{Experiments}
\label{sec:exps}

\noindent
\textbf{Dataset and Metrics.}
FDDiff is evaluated on face SR and general SR.
For large-scale face images, following \cite{SR32023,DiWa2023,pulse2020}, FDDiff is trained on $7\times 10^4$ images from Flickr-Face-HQ(FFHQ) \cite{FFHQ2019} and evaluated on CelebA-HQ \cite{CelebAHQ2018}.
Following \cite{SR32023,IDM2023,DiWa2023}, super resolution experiments for $16\times16\rightarrow128\times128$ with upscaling ratios $8\times$ are conducted.
For general images, FDDiff is evaluated on DIV2K \cite{DIV2K2017} benchmark for $4\times$ SR task.
Peak Signal-to-Noise Ratio (PSNR) and Structural Similarity Index Measure (SSIM) are used to evaluate SR from numerical and perceptual views.

\noindent
\textbf{Implementation Details.}
FDDiff was trained from scratch on RTX3090s for 1M steps with batch sizes of 64. 
AdamW optimizer with $5\times 10^{-4}$ learning rate  is adopted.
We utilize U-Net architecture for FDDiff.
The channel dimensions for $8\times$ SR with $p=3$ of $p+1=4$ encoder blocks are $[64,128,256,128]$, and the ones for $4\times$ SR are $[128,256,128]$. 
Inference time per step is 17.14 ms.

\noindent
\textbf{Large-scale Face Super-Resolution.}
As shown in Table \ref{tab.8xfaceSR}, FDDiff outperforms previous generative methods both in terms of PSNR and SSIM.
Compared with up-to-date IDM, FDDiff obtains better results with 0.51dB higher in PSNR.
For qualitative comparison in Fig. \ref{fig.face}, FDDiff generates more faithful facial details, e.g., eyes, teeth, and noses.

\noindent
\textbf{General Image Super-Resolution.}
As shown in Table \ref{tab.4xDIV2K}, FDDiff achieves competitive numerical SR quality and outstanding perceptual SR quality, outperforming previous by at least 0.14dB for PSNR and $11.4\%$ for SSIM.
It reveals that FDDiff prioritizes human perceptual quality.

\noindent
\textbf{Real-world Super-Resolution.}
Table \ref{tab.realsr} further shows real-world SR experiment, where FDDiff has shown comparable PSNR and SSIM, trained by pipeline of Real-ESRGAN\cite{esrgan2018}. 

\noindent
\textbf{Ablation Studies.}
In Table \ref{tab.ExpAblationFDDiff}, we evaluate different scale settings, target numbers, and targets of diffusion process for $8\times$ face SR on CelebA-HQ.
Model 1 is adapted to single scale and only predicts noise $\boldsymbol{\epsilon}_t$, similar to prior works.
Model 2 to 3 progressively predict finer-grained frequency targets, while Model 4 remove diffusion process.
Model 3 in the gray row, i.e., FDDiff, has achieved higher SR quality, owing to its fine-grained frequency partition.
Model 4 obtains lower PSNR and SSIM, indicating the effectiveness of diffusion process.

For infrence efficiency, FDDiff can achieve decent performance with only $64$ timesteps, as shown in Fig. \ref{Fig.ExpTimestep}.

\section{Conclusion}
\label{sec:conclusion}
In this paper, we propose FDDiff framework that decomposes super-resolution tasks into progressive, finer-grained denoising and frequency-refinement steps. 
Utilizing wavelet packet-based frequency chain to generate multiscale targets, FDDiff designs a non-Markovian diffusion process to denoise and complement missing details simultaneously.
Comprehensive evaluations demonstrate that FDDiff can effectively reduce artifacts and achieve higher super-resolution accuracy.

\vfill\pagebreak
\bibliographystyle{IEEEbib}
\bibliography{refs}

\end{document}